\documentclass[10pt, a4paper]{article}
\usepackage{lrec}
\usepackage{multibib}
\usepackage{graphicx}
\usepackage{tabularx}
\usepackage{soul}

\usepackage{epstopdf}
\usepackage[latin1]{inputenc}

\usepackage{hyperref}
\usepackage{xstring}

\usepackage{booktabs,multirow,threeparttable,fancyhdr}
\renewcommand{\thesection}{\arabic{section}}
\renewcommand{\thesubsection}{\arabic{section}.\arabic{subsection}}

\title{A Large Self-Annotated Corpus for Sarcasm}

\name{Mikhail Khodak, Nikunj Saunshi, Kiran Vodrahalli}

\address{Computer Science Department, Princeton University \\
         35 Olden St., Princeton, New Jersey 08540 \\
         {\tt \{mkhodak,nsaunshi,knv\}@cs.princeton.edu}}

\abstract{
We introduce the Self-Annotated Reddit Corpus ({\bf SARC}), a large corpus for sarcasm research and for training and evaluating systems for sarcasm detection.
The corpus has 1.3 million sarcastic statements --- 10 times more than any previous dataset --- and many times more instances of non-sarcastic statements, allowing for learning in both balanced and unbalanced label regimes.
Each statement is furthermore {\em self-annotated} --- sarcasm is labeled by the author, not an independent annotator --- and provided with user, topic, and conversation context. 
We evaluate the corpus for accuracy, construct benchmarks for sarcasm detection, and evaluate baseline methods.
\\\newline  }

\begin{document}

\maketitleabstract


\section{Introduction}
\label{sec:intro}

Sarcasm detection is an important component in many natural language processing (NLP) systems, directly relevant to natural language understanding, dialogue systems, and text mining.
However, detecting sarcasm is difficult because it occurs infrequently and is difficult for even humans to discern \cite{Wallace:14}.
Despite these properties, existing datasets either have {\em balanced labels} --- data with roughly the same number of examples of each label \cite{Gonzalez:11,Bamman:15,Joshi:15,Amir:16,Oraby:16} --- or use humans to annotate sarcastic statements \cite{Riloff:13,Swanson:14,Wallace:15}.

In this work, we make available the first corpus\footnote{\url{http://nlp.cs.princeton.edu/SARC/}} for sarcasm detection that has both unbalanced and self-annotated labels and does not consist of short text snippets from Twitter\footnote{\url{https://www.twitter.com}}.
With more than a million examples of sarcastic statements, each provided with author, topic, and contex information, the dataset exceeds all previous sarcasm corpora by an order of magnitude in size.
This is possible due to the comment structure of the social media site Reddit\footnote{\url{https://www.reddit.com}} as well as its frequently-used and standardized annotation for sarcasm.

Following a discussion of corpus construction and relevant statistics in Section~\ref{sec:details}, we discuss the quality of this dataset compared to alternative sources in Section~\ref{sec:eval}, manually evaluating our corpus for noise.
Then in Section~\ref{sec:detect} we use our dataset to construct suitable benchmarks for sarcasm detection systems and examine the performance of simple baseline methods and human evaluators on these subsets.


\section{Related Work}
\label{sec:related}
Since our main contribution is a corpus and not a method for sarcasm detection, we point the reader to a recent survey by \newcite{Joshi:16} that discusses many interesting efforts in this area.
Note that many of the works the authors mention will be discussed by us in this section, with many papers using their own datasets and illustrating the need for common evaluation baselines.

Sarcasm datasets can largely be distinguished by the sources used to get sarcastic and non-sarcastic statements, the amount of human annotation, and whether the dataset is balanced or unbalanced.
Reddit has been used before, notably by \newcite{Wallace:15}; 
while the authors allow unabalanced labeling, they do not exploit the possibility of using self-annotation and generate around 10,000 human-labeled sentences.
Twitter is a frequent source due to the self-annotation provided by hashtags such as \#sarcasm, \#notsarcasm, and \#irony \cite{Reyes:13,Bamman:15,Joshi:15}.
As discussed in Section \ref{subsec:twitter}, its abbreviated language and other properties make Twitter a less attractive source for annotated comments.
However, it is by far the largest raw source of data for this purpose and has led to some large unbalanced corpora in previous efforts \cite{Riloff:13,Ptacek:14}.
A further source of comments is the Internet Argument Corpus (IAC) \cite{Walker:12}, a scraped corpus of Internet discussions that can be further annotated for sarcasm by humans or by machine learning; 
this is done by \newcite{Lukin:13} and \newcite{Oraby:16}, in both cases resulting in around 10,000 labeled statements.


\section{Corpus Details}
\label{sec:details}
\subsection{Reddit Structure and Annotation}
\label{subsec:reddit}

Reddit is a social media site in which users communicate by commenting on {\em submissions}, which are titled posts consisting of embedded media, external links, and/or text, that are posted on topic-specific forums known as {\em subreddits}; examples of subreddits include \texttt{funny}, \texttt{pics}, and \texttt{science}.
Users comment on submissions and on other comments, resulting in tree-like conversation structure such that each comment has a parent comment. 
We refer to {\em elements} as any nodes in the tree of a Reddit link (i.e., comments or submissions). 

Reddit users have adopted a common method for sarcasm annotation consisting of adding the marker ``/s" to the end of sarcastic statements; 
this originates from the HTML text delineation \texttt{<sarcasm>\dots</sarcasm>}.
As with Twitter hashtags, using these markers as indicators of sarcasm is noisy  \cite{Bamman:15}, especially since many users do not use the marker, do not know about it, or only use it where sarcastic intent is not otherwise obvious.
We discuss the extent of this noise in Section \ref{subsec:manual}.

\subsection{Constructing SARC}
\label{subsec:sarc}

\begin{figure}
	\centering
	\includegraphics[scale=0.66]{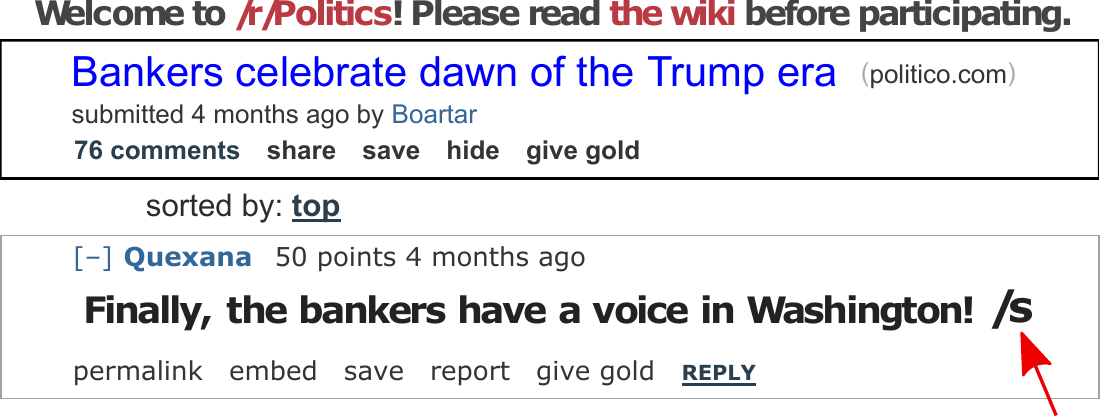}
	\caption{\label{fig:example}
A Reddit submission and one of its comments.
Note the conventional annotation ``/s" indicating sarcasm.}
\end{figure}

Reddit comments from December 2005 have been made available due to web-scraping \footnote{\url{http://files.pushshift.io/reddit}};
we construct our dataset as a subset of comments from January 2009-April 2017, comprising the vast majority of comments and excluding noisy data from earlier years.
For each comment we provide a sarcasm label, author, the subreddit it appeared in, the comment score as voted on by users, the date of the comment, and identifiers linking back to the original dataset of all comments.

To reduce noise, we use several filters to remove noisy and uninformative comments.
Many of these are standard preprocessing steps such as excluding URLs and limiting characters to be ASCII.
To handle Reddit data, we also exclude comments that are descendants of sarcastic comments in the conversation tree, as annotation in such cases is extremely noisy, with authors agreeing or disagreeing with the previously expressed sarcasm with their own sarcasm but often with no marking.

Our raw corpus consists of three files:
\begin{enumerate}
	\item An array in CSV format containing 533 million comments, of which around 1.3 million are sarcastic. This file only contains those comments whose authors know about the standard sarcasm annotation; this is determined by whether they have used the annotation in the same month as the comment was made or earlier. This limitation is added in order to reduce false negatives due to authors not annotating their sarcasm. Each row also contains the parent comment.
	\item A hashtable in JSON format containing all comments and posts in the conversation thread of a sarcastic comment as well as all siblings of sarcastic comments.
	\item An array in CSV format, with each row containing a sequence of comments leading up to a sarcastic comment, the (sarcastic and non-sarcastic) responses to the last element in that sequence, and the labels of those responses. Each element is given as a key to the previous file.
\end{enumerate}

This raw corpus is very large and suitable for both large-scale machine learning and statistical analysis as well for deriving smaller benchmark tasks for evaluating sarcasm detection systems.
These benchmarks, whether in the balanced or unbalanced regimes, require further subsampling of the corpus and an approach for dealing with noisy data in the face of sparse signals.
We specify and evaluate an approach for doing so in Section~\ref{subsec:task}, followed by the evaluation of learning algorithms on the output.

\begin{table}[!t]
\centering
\begin{threeparttable}
\begin{tabular}{@{}llcc@{}}
Corpus & Dataset & Sarcastic & Total \\
\toprule
\multirow{2}*{IAC}
& Joshi et al. `15 & 751 & 1502 \\
& Oraby et al. `16 & 4.7K & 9.4K \\
\midrule
\multirow{5}*{Twitter}
& Joshi et al. `16 & 4.2K & 5.2K \\
& Bamman \& Smith `15 & 9.7K & 19.5K \\
& Reyes et al. `13 & 10K & 40K \\
& Riloff et al. `13 & 35K & 175K \\
& Pt\'{a}\u{c}ek et al. `13 & 130K & 780K \\
\midrule
\multirow{2}*{Reddit}
& Wallace et al. `15 & 753 & 14124 \\
& {\bf SARC} & {\bf 1.34M} & {\bf 533M} \\
\bottomrule
\end{tabular}
\end{threeparttable}
\caption{\label{tbl:stats} 
SARC compared with previous sarcasm corpora. In addition to a million sarcastic comments our dataset also provides many millions more non-sarcastic statements by the same authors.}
\end{table}


\section{Corpus Evaluation}
\label{sec:eval}

There are three major metrics of interest for evaluating our corpora: (1) size, (2) the proportion of sarcastic to non-sarcastic comments, and (3) the rate of false positives and false negatives.
Of interest is also the quality of the text in the corpus and its applicability to other NLP tasks.
Thus in this section we evaluate error in the raw corpus and provide comparison with other corpora used to construct sarcasm datasets. We also discuss the potential limitations of our approach.

\subsection{Manual Evaluation}
\label{subsec:manual}

To investigate the noisiness of using Reddit as a source of self-annotated sarcasm we estimate the proportion of false positives and false negatives induced by our filtering.
This is done by manually checking a random subset of 500 comments from SARC tagged as sarcastic and 500 tagged as non-sarcastic, with full access to the comment's context.
A comment was determined to be false positive if ``/s" tag was not an annotation but part of the sentence and a false negative if the comment author was clearly being sarcastic to the human rater.
This procedure yielded a false positive rate of 1.0\% and a false negative rate of 2.0\%.
Although the false positive rate is reasonable, the false negative rate is significant compared to the sarcasm proportion (0.25\%), indicating large variation in the working definition of sarcasm and the need for methods that can handle noisy data in the unbalanced setting.
In the balanced setting this is still a fairly small amount of noise.  

\subsection{Comparison with other Sources}
\label{subsec:twitter}

As noted before, Twitter has been the most common source for sarcasm in previous corpora;
this is likely due to the explicit annotation provided by its hashtags.
However, using Reddit as a source of sarcastic comments holds many research advantages.
Unlike Reddit comments, which are not constrained by length and contain fewer hashtags, tweets are written in abbreviated English.
Hashtagged tokens are also frequently used as a part of the statement itself (e.g. ``that was \#sarcasm''), blurring the line between text and annotation; on Reddit ``/s" is generally only used as something other than annotation when its use as an annotation is being referred to (e.g. ``you forgot the /s").
The full conversation context is also much easier to provide on Reddit due to the shallow tree structure of an individual post and its comments.

Furthermore, from a subsample of Twitter and Reddit data from July 2014 we determined that a vastly smaller percentage (.002\% vs. .927\%) of Twitter authors make use of sarcasm annotation (\#sarcasm, \#sarcastic, or \#sarcastictweet).
We hypothesize that Reddit users require sarcastic annotation more frequently and in a more standardized form because they are largely anonymous and so cannot rely on a shared context to communicate sarcasm.
Finally, Reddit also benefits from having subreddits, which enable featurization and data exploration based on an explicit topic assignment.

\begin{figure}
	\centering
	\includegraphics[scale=0.36]{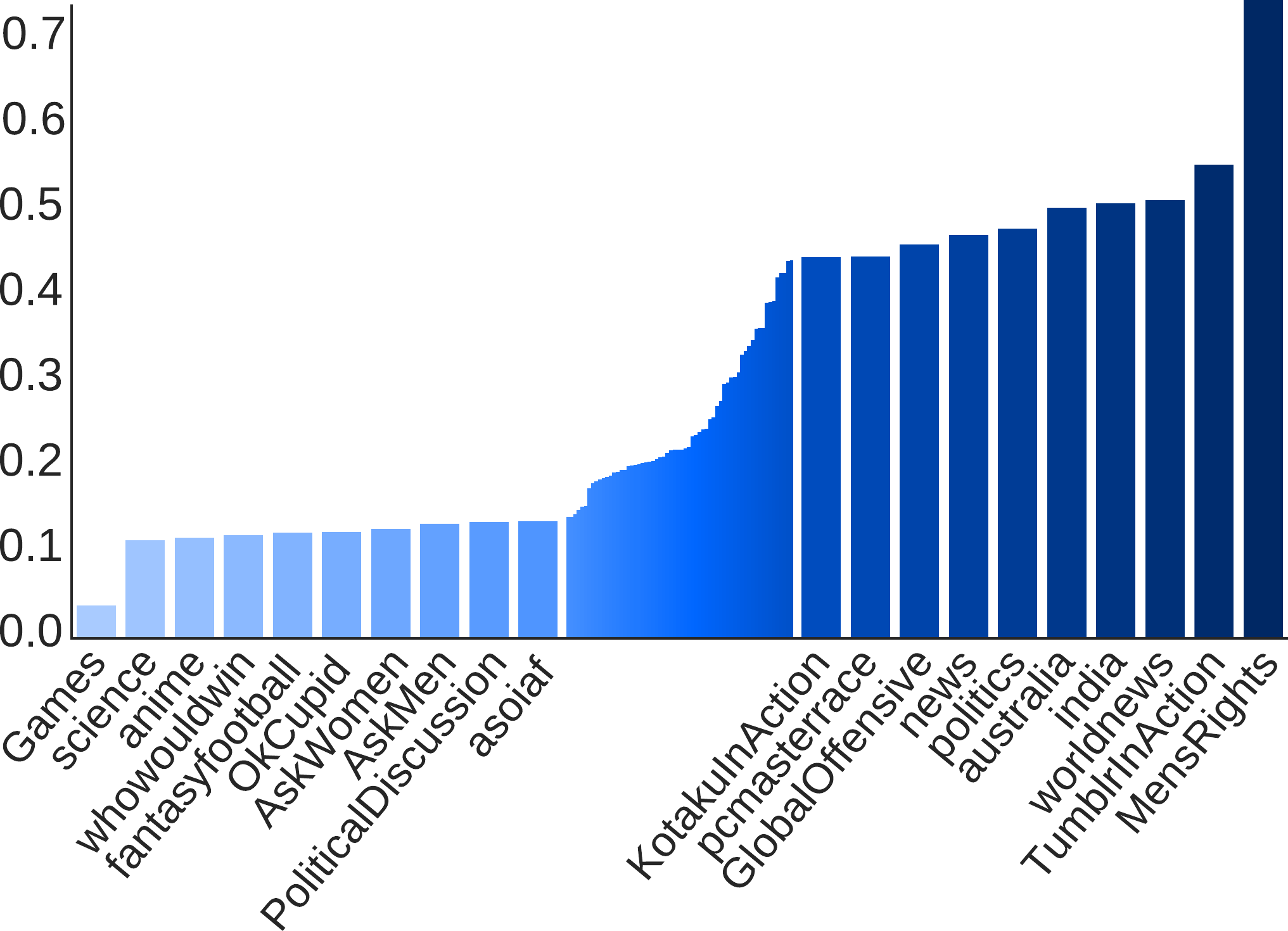}
	\caption{\label{fig:subreddit_sarcasm} Sarcasm percentage for subreddits with more than a million comments in SARC. 
Well-moderated and special-interest forums such as \texttt{science} and \texttt{asoiaf} (referring to fantasy series \textit{A Song of Ice and Fire}) have less sarcasm than controversial and less-moderated subreddits.}
\end{figure}

The Internet Argument Corpus (IAC) has also been used as a source of sarcastic comments \cite{Walker:12}.
The corpus developers found 12\% of examples in the IAC to be sarcastic, which is a much nicer class proportion for sarcasm detection than ours.
As the Reddit data consists of arbitrary conversations, not just arguments, it is not surprising that our sarcasm percentage is much smaller, even when accounting for false negatives;
this property also makes our dataset more realistic.
Unlike Reddit and Twitter, the IAC also requires manual annotation of sarcasm.


\subsection{Limitations of Our Approach}
\label{subsec:limits}

There are a few noteworthy limitations to our method of collecting a self-annotated sarcasm dataset. Despite our efforts to filter noisy ``/s'' labels, there remain instances where no simple rule reliably eliminates incorrect labels. We describe the difficulties for both false positives and false negatives:
\begin{itemize}

\item \textbf{False positives} are instances where a comment is incorrectly labeled as sarcastic due to the presence of a ``/s'' tag. This case only occurs when a ``/s'' tag appears in the comment with a meaning different from indicating sarcasm. As previously noted, this possibility is more likely to occur if a user is unaware of the ``/s'' notation. Similarly, if ``/s'' is used to \textit{refer to the convention of its use as an annotation}, the naive approach of merely detecting the ``/s'' string also fails. Finally, it is possible that ``/s'' has other connotations: For instance, in HTML, \texttt{<s>\dots</s>} denotes a strike-through. Therefore a subreddit focusing on the discussion of web programming, for example, might include instances where ``/s'' is used with a different meaning.

To combat the first issue, we restrict to users aware of the sarcasm notation by ensuring they have used ``/s'' previously. This filter helps ensure that the users are aware of the semantic usage of ``/s''. For the second case, we only keep comments which have the ``/s''  at the \textit{end} of the comment. All comments we inspected which terminated in ``/s'' used the annotation to indicate sarcasm. The third case amounts to solving word sense disambiguation, and we did not find a universally simple approach to reduce noise of this form. However, it is possible to reduce the likelihood of this form of sense mismatch by restricting to subreddits which are known to not have alternate senses for ``/s'' (e.g., politics). 

\item \textbf{False negatives} are instances where a comment is sarcastic, but not annotated with a ``/s''. False negatives are harder to detect than false positives since the portion of comments which have no sarcasm annotation is much larger than the portion that do. There are two primary ways a false negative can arise: Either a user does not know of the ``/s'' convention, or the user believes their use of sarcasm is obvious enough to warrant not including the tag. Notably, such a belief depends on what community the user is communicating in, who the user is communicating with (another user they routinely have arguments with, or a stranger), and also on prior comments on the thread. As noted previously, a comment which is sarcastic often spawns a chain of subsequent comments which are all sarcastic, but which lack the ``/s'' symbol. In short, context matters a lot for determining whether or not a sarcastic comment is obvious. 

The first issue is solved by our first filter. The second issue is difficult to address, and remains a limitation of our approach. We avoid the particular case of sarcastic comment chains by discarding the child comments of sarcastic comments in a thread.

\end{itemize}

All of our filters are validated by manual evaluation of the false positive and false negative rates as described in Section~\ref{subsec:manual}, which improved considerably after implementing the filters. Our manual evaluation approach has one central limitation: Though we provide local context to the human annotators, if the ability to distinguish the sarcastic intent of a comment relies on knowledge of, for instance, the commenter's comment history or relevant news, then human annotators may not perform well. We tried to resolve this issue by using a voting scheme which required several humans to agree about whether or not a comment with its context was sarcastic.


\section{Benchmarks for Sarcasm Detection}
\label{sec:detect}

\begin{table}[!t]
	\centering
	\begin{threeparttable}
		\begin{tabular}{@{}lccc@{}}
			Method & $\textrm{all-bal}^\ast$ & pol-bal & pol-unbal$^\dagger$ \\
			\toprule
			Bag-of-Words & 73.2 & 75.9 & 27.0\\
			Bag-of-Bigrams & 75.8 & 76.5 & 24.9\\
			Sentence Embedding & 71.0 & 76.0 & 26.7\\
			Human (Average) & 81.6 & 83.0 & -\\
			Human (Majority) & 92.0 & 85.0 & -\\
			Random & 50.0 & 50.0 & 10.2\\
			\bottomrule
		\end{tabular}
	\begin{tablenotes}
		\item[$\ast$] Used only features appearing at least 5 times in the corpus for Bag-of-Words and Bag-of-Bigrams.
		\item[$\dagger$] Used only features appearing at least 100 times in the corpus for Bag-of-Words and Bag-of-Bigrams. Measured as average $F_1$- score scaled by 100
	\end{tablenotes}
	\end{threeparttable}
		\caption{\label{tbl:performance}Accuracy percentage of baseline methods for sarcasm detection compared to human performance. Tests conducted in the balanced regime for the all subreddits task and balanced and unbalanced regimes for the politics subreddit task}
\end{table}

A direct application of our corpus is for training and evaluating sarcasm detection systems.
Thus we use the raw corpus described in Section~\ref{sec:details} to construct several useful benchmarks for the task of classifying statements as sarcastic or non-sarcastic.
All benchmarks provide the full conversation thread leading up to the target statements to the learning algorithm, along with comment metadata.
Following their specification we consider a few context-free baseline methods depending only on linear classification over simple featurizations.
Code to reproduce our results is provided at \url{https://github.com/NLPrinceton/SARC}.

\subsection{Evaluation Task}
\label{subsec:task}

In the most general case, we use the provided raw files to construct datapoints for systems to learn the following task: given a post and a sequence of comments, determine which comments among the responses to the last comment in the sequence are sarcastic.
Thus each datapoint consists of a conversation thread followed by a series of responses and sarcasm labels.
Performance on this task is measured by average precision, recall and $F_1$ scores.

Before constructing this subcorpus we first remove from consideration all comments that are not complete sentences and not between 2 and 50 tokens long, allowing for cleaner comments in the evaluation. 
Although the responses are still largely non-sarcastic, the proportion of sarcastic comments is much greater here as each datapoint must correspond to a thread where at least one sarcastic annotation occurred.
In total we construct 8.44 million sequences, with the average proportion of sarcastic responses being 28.1\%.

\subsubsection{Balanced Labels}
We construct a balanced learning task by taking only one sarcastic and one non-sarcastic response from each set of responses to a comment sequence.
The task then becomes one of picking which of two statements that share a context is sarcastic, with performance measured by accuracy.

While having only posts with at least one sarcastic response is useful, it also increases the false negative rate as comments warranting a sarcastic response often draw other sarcastic statements that are similar in content to the labeled sarcastic responses, but which themselves may not be labeled.
Thus to reduce this issue when picking the non-sarcastic statement, we featurize all statements using the normalized sum of Common Crawl GloVe embeddings of the words and pick from only those non-sarcastic statements that have similarity $\le0.95$ with the sarcastic statement \cite{Pennington:14}. 

\subsubsection{Politics}
The difficulty of detecting sarcasm rests not only on the need to understand the context of previous statements but also on understanding background information on the topic being discussed.
Even humans will struggle with sarcastic comments drawn from unfamiliar topics, for instance, obscure hobbies or art forms.
Thus we also test human and machine performance on comments drawn solely from the \texttt{politics} subreddit, a topic for which all evaluators had sufficient background information.
This subsample contains 17 thousand sequences, with the average proportion of sarcastic responses being 23.2\%.

\begin{figure}
	\centering
	\includegraphics[scale=0.48]{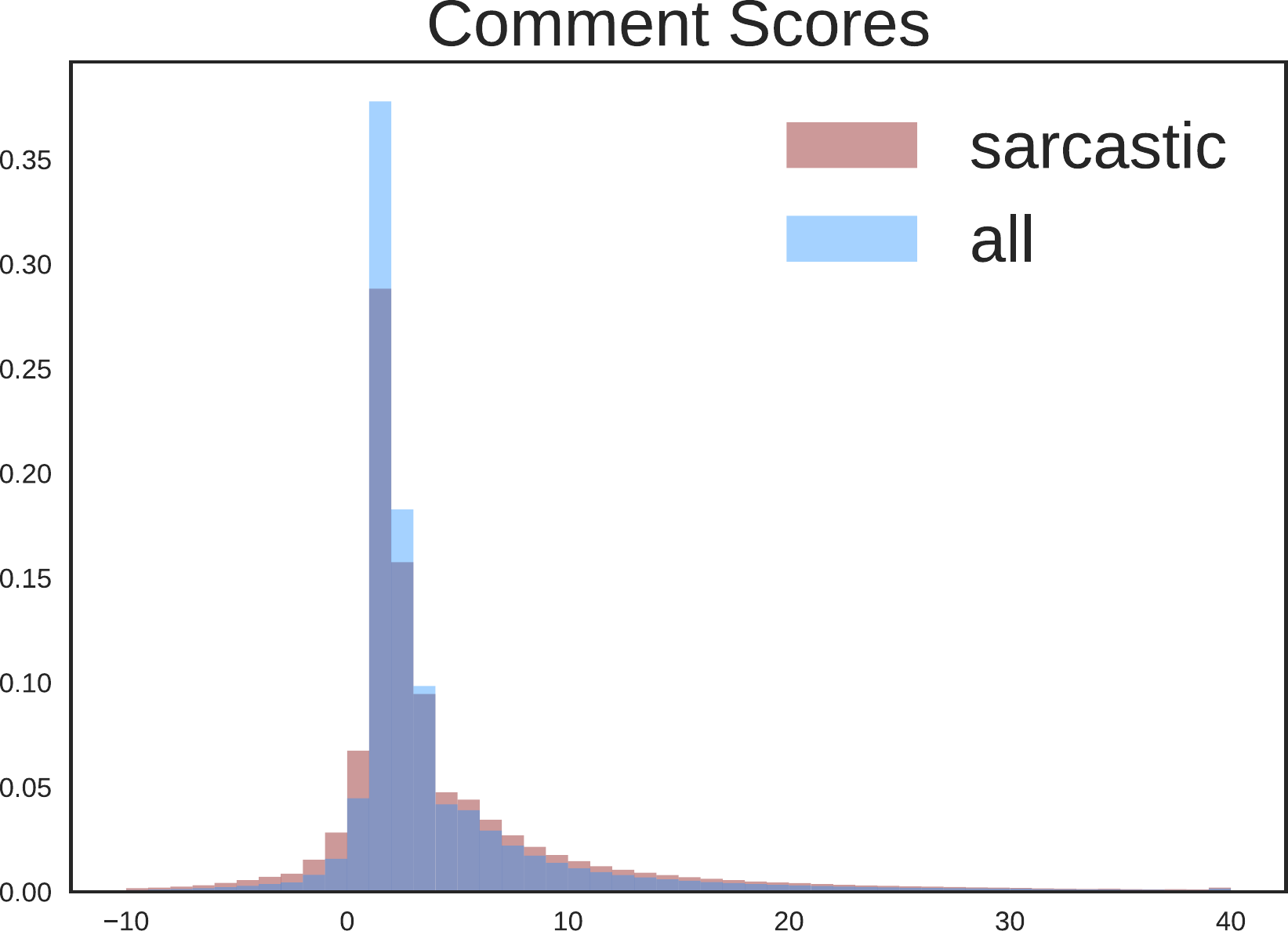}
	\caption{\label{fig:scores} Score distributions of sarcastic and non-sarcastic comments in the raw SARC dataset.}
\end{figure}

\begin{figure}
	\centering
	\includegraphics[scale=0.48]{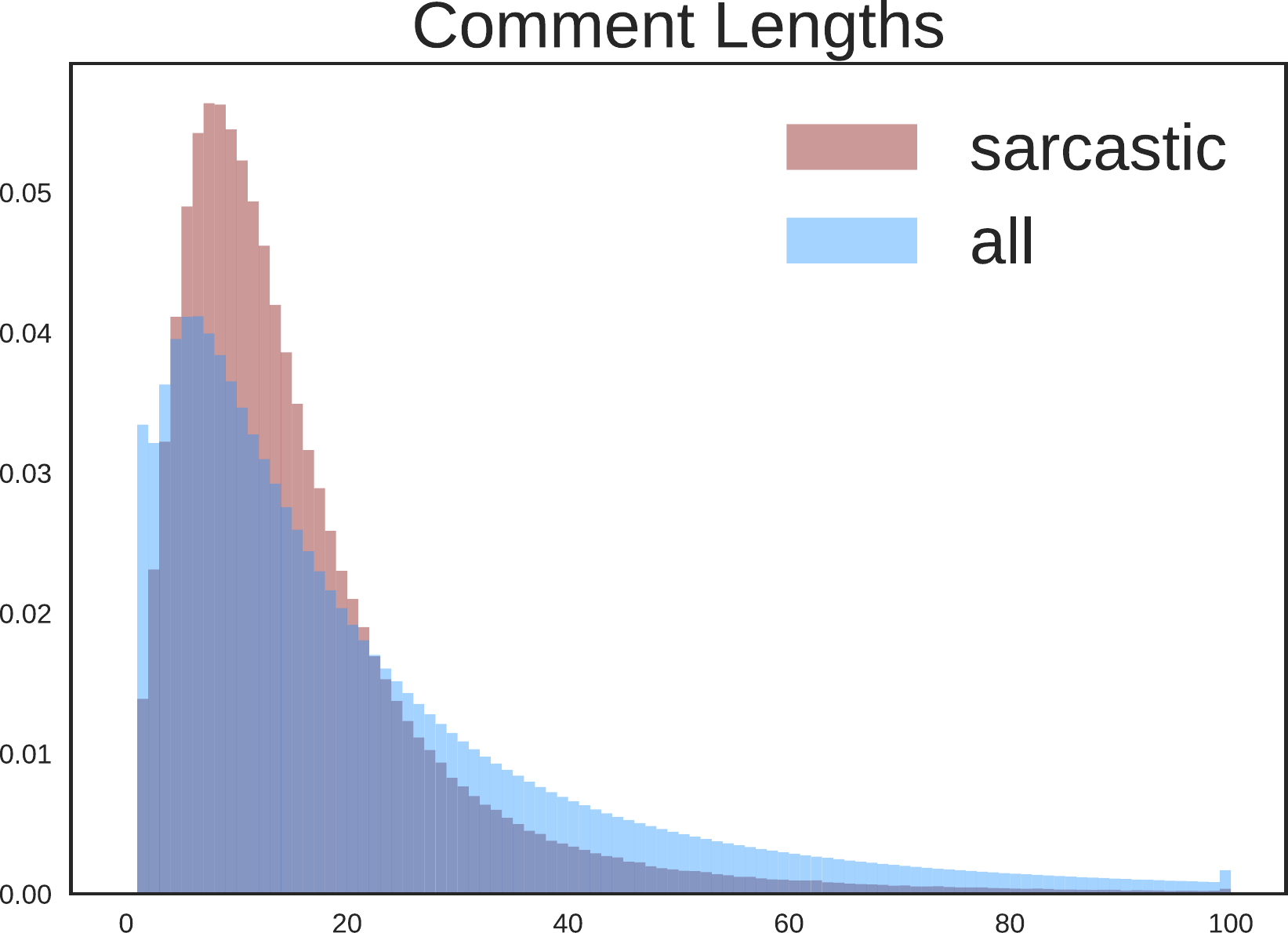}
	\caption{\label{fig:lengths} Length distributions of sarcastic and non-sarcastic comments in the raw SARC dataset.}
\end{figure}

\subsection{Methods}
\label{subsec:methods}
For the case of balanced labels, a simple, no-context baseline method for the above task is to featurize the two responses and to train a logistic regression classifier to distinguish between the sarcastic and non-sarcastic response as separate classes.
On the testing set, we pick the response with the highest probability of being labeled sarcastic as the sarcastic one.
We split both datasets we test on 80\%-20\% between train-test subsets and report the results of the following three approaches in Table~\ref{tbl:performance}.

\subsubsection{Bag-of-$n$-Grams}
\label{subsec:bong}

The Bag-of-$n$-Grams representation consists of using a document's $n$-gram counts as features in a vector.
We test two variants, the Bag-of-Words and the Bag-of-Bigrams.
For the subsample containing all subreddits we use only those features that occur at least 5 times in the training comments.
We considered including other comment features, such as comment length and scores, but empirical results and the distributions of these features (see Figures~\ref{fig:scores} and~\ref{fig:lengths}) indicate that they are not particularly informative.
More sophisticated featurization, such as the noun-phrase and feature interaction indicators proposed by \cite{Wallace:14}, is left to future work.

\subsubsection{Sentence Embeddings}
\label{subsec:sif}

Given a document, taking the elementwise sum of embeddings of its words provides a simple low-dimensional document representation. This particular technique of constructing word sequence featurizations has been previously studied and established as a strong baseline for a multitude of supervised NLP prediction tasks \cite{Arora:17}. 
We use 1600-dimensional GloVe representations trained on the Amazon product corpus, which is used instead of Common Crawl because of the semantic closeness between sentiment and sarcasm \cite{McAuley:15}.

\subsubsection{Human}
\label{subsec:human}

Human sarcasm detection performance was measured by giving 5 human evaluators 100 samples and asking them to perform the same task as the algorithm: determining which of two statements is sarcastic. We provide links to the evaluation survey for the full corpus\footnote{\texttt{www.surveygizmo.com/s3/3878814/SARCmain}} as well as the politics subreddit\footnote{\texttt{www.surveygizmo.com/s3/3878798/SARCpol}}.
The full context was provided, and the final human classifier was taken as the majority vote of all 5 evaluators.

\subsubsection{Random}
\label{subsec:random}

We use a simple baseline where all responses are labeled sarcastic randomly and independently with a fixed probability. This probability is chosen as the average fraction of responses that are sarcastic in the training set.

\subsection{Results}
\label{subsec:results}

\begin{table}
	\centering
	\begin{tabular}{c c c c}
		Pos. $n$-Grams & Weights & Neg. $n$-Grams & Weights\\
		\toprule
		obviously & 1.79 & :) & -1.37\\
		clearly & 1.66 & lmao & -1.27\\
		so fun & 1.49 & :( & -1.17\\
		totally & 1.39 & :/ & -1.17\\
		good thing & 1.35 & , but & -1.10\\
		shocked & 1.32 & lol & -1.00\\
		shocking & 1.23 & the original & -0.98\\
		m sure & 1.15 & wat & -0.97\\
		omg & 1.13 & why & -0.96\\
		how dare & 1.13 & oh god & -0.95\\
	\end{tabular}
	\caption{Most positive and negative $n$-grams based on weights assigned by the Bag-of-Bigrams classifier. Positive (negative) weight for an $n$-gram implies it is a strong indicator for the comment being sarcastic (non-sarcastic). The weights indicate that positive $n$-grams are more important for linear classification of sarcasm.}
\end{table}

\subsubsection{Baselines}
The baselines in Table~\ref{tbl:performance} perform reasonably well and much better than the random baseline, but none of them match human performance on either dataset. There is clear scope for improvement for machine learning methods, starting with the use of context provided to make better decisions about sarcasm. As evident in Table~\ref{tbl:performance}, Bag-of-Word and Bag-of-Bigram representations perform better than sentence embeddings; however, distributed representations may be necessary for incorporating context in future methods.

\subsubsection{Human}
As expected, human evaluators performed significantly better, both as a majority and on-average, than the baseline methods. There was significant but not perfect agreement among annotators: on the main dataset the Fleiss kappa score \cite{Fleiss:71} was 0.5, indicating moderate agreement, while on the politics subsample it was 0.67, indicating substantial agreement. Interestingly, while individually human performance was worse on average on sequences drawn from all subreddits than on the politics subsample, taking a majority vote among humans led to much better performance in the former case. This performance boost indicates that while individuals may not have enough context for all topics of discussion on Reddit, in aggregate there is enough information to do well, even surpassing the performance on a well-known topic such as politics.


\section{Conclusion}
\label{sec:conclusion}

We introduce a large sarcasm dataset based on self-annotated Reddit comments.
Both the raw data and evaluation subsamples are made freely available, with the former having over 1 million sarcastic sentences, larger than any existing dataset.
We evaluate the baseline performance of simple machine learning methods and compare them with human performance.
We hope that future users of this dataset will improve upon these benchmarks and find new ways of utilizing the large quantities of self-annotated information we provide.

\section{Acknowledgements}
We would like to acknowledge Angel Chang and Christiane Fellbaum for helpful discussion and suggestions.

\section{Bibliographical References}
\bibliographystyle{lrec}
\bibliography{reference}

%

\end{document}